\documentclass[conference]{IEEEtran}
\IEEEoverridecommandlockouts

\usepackage{cite}
\usepackage{amsmath,amssymb,amsfonts}
\usepackage{graphicx} 
\usepackage{multirow} 
\usepackage{booktabs} 
\usepackage{hyperref}
\usepackage{algorithmic}
\usepackage{graphicx}
\usepackage{textcomp}
\usepackage{xcolor}
\usepackage{tikz}

\def\BibTeX{{\rm B\kern-.05em{\sc i\kern-.025em b}\kern-.08em
    T\kern-.1667em\lower.7ex\hbox{E}\kern-.125emX}}

\begin{document}

\title{Contrastive Continual Learning for Model Adaptability in Internet of Things}

\author{
\IEEEauthorblockN{Ajesh Koyatan Chathoth}
\IEEEauthorblockA{}
}

\maketitle

\begin{abstract}
Internet of Things (IoT) deployments operate in nonstationary, dynamic environments where factors such as sensor drift, evolving user behavior, and heterogeneous user privacy requirements can affect application utility. Continual learning (CL) addresses this by adapting models over time without catastrophic forgetting. Meanwhile, contrastive learning has emerged as a powerful representation-learning paradigm that improves robustness and sample efficiency in a self-supervised manner. This paper reviews the usage of \emph{contrastive continual learning} (CCL) for IoT, connecting algorithmic design (replay, regularization, distillation, prompts) with IoT system realities (TinyML constraints, intermittent connectivity, privacy). We present a unifying problem formulation, derive common objectives that blend contrastive and distillation losses, propose an IoT-oriented reference architecture for on-device, edge, and cloud-based CCL, and provide guidance on evaluation protocols and metrics. Finally, we highlight open unique challenges with respect to the IoT domain, such as spanning tabular and streaming IoT data, concept drift, federated settings, and energy-aware training.
\end{abstract}


\section{Introduction}
Internet of Things (IoT) analytics increasingly relies on machine learning models deployed across sensors, gateways, edge devices, and the cloud. In practice, IoT data streams are \emph{non-stationary}: distributions shift due to seasonal effects, privacy preferences, device aging, firmware updates, and changing user routines~\cite{li2017weather, chathoth2025dynamic}. IoT devices generate large volumes of sensor data that can be leveraged for applications such as network anomaly detection, surveillance, and smart home control~\cite{chathoth2025pcap}. However, training once (offline) and deploying forever does not work in such evolving scenarios and leads to performance decay. Continual learning (CL) tackles this by learning a sequence of tasks or data segments while retaining prior knowledge, thereby mitigating catastrophic forgetting \cite{kirkpatrick2017ewc,rebuffi2017icarl, chathoth2021federated, chathoth2022differentially}.

Machine learning based applications are enormous, and their possibilities have further magnified due to the availability of large data generated by IoT devices and recent advancements in large language models~\cite{xu2014iiot,peng2025log}.
Contrastive learning is a powerful approach in machine learning that focuses on the relationships between data points and represents them in a latent space for each classification task, thereby enhancing model performance with limited labeled data during training~\cite{khosla2020supcon, khosla2020supcon}. By contrasting similar and dissimilar samples, the model learns to identify features that distinguish one class from another. This method is particularly effective for tasks such as image recognition and natural language processing, where input variability can be substantial. By maximizing agreement among similar examples and minimizing it among dissimilar ones, contrastive learning fosters a deeper understanding of the underlying data structure, thereby improving accuracy and robustness across various applications~\cite{hu2024comprehensive}.
At the same time, contrastive representation learning (e.g., SimCLR, MoCo, BYOL) has improved robustness, privacy, and transfer by learning invariances via augmented views and similarity objectives \cite{chen2020simclr,he2020moco,grill2020byol, chathoth2025dynamic}. 
Contrastive learning is also used in IoT use cases for privacy-preserving techniques~\cite{chathoth2025dynamic}.

Recent work shows that blending contrastive losses with CL strategies can preserve \textit{useful representations over time}, thereby preserving the utility. One such technique is Co$^2$L, which combines supervised contrastive learning with self-distillation and replay to reduce forgetting \cite{cha2021co2l}.Co$^2$L employs a rehearsal-based continual learning algorithm that focuses on continually learning and maintaining transferable representations.
IoT devices are resource-constrained and necessitate security and privacy-preserving techniques that account for their memory and power consumption~\cite{melnyk2025hardware}.
Prior work has focused on the vision domain, and these techniques may not be directly applicable to resource-constrained IoT devices. In contrast, this paper examines \emph{contrastive continual learning (CCL) for IoT}, where (i) labels are scarce or delayed, (ii) data is heterogeneous and often tabular/time-series rather than images, and (iii) computation, memory, energy, and bandwidth constraints are central (TinyML/edge) \cite{warden2019tinyml,lourenco2025ondevice}. We aim to connect algorithms and systems, providing an implementable blueprint and evaluation guidance.
Our main contributions are:
%\textbf{Contributions.} We:
\begin{itemize}
  \item Propose an IoT-oriented architecture for on-device, edge, and cloud CCL with practical design schemes.
  \item Formalize IoT contrastive continual learning objectives and training methodologies (task-incremental and stream-incremental).
  \item Summarize core method families: replay with contrastive, distillation-based CCL, and its theory and guarantees.
  \item Recommend evaluation protocols and metrics tailored to IoT constraints and concept drift.
  \item Identify open problems for tabular and time-series CCL, federated drift, privacy, and energy-aware learning.
\end{itemize}

\section{Background and Related Work}
\subsection{Continual Learning}
CL methods commonly fall into: (i) \emph{regularization} (protect important parameters), (ii) \emph{replay} (store or generate past samples), and (iii) \emph{architectural} expansion (add capacity). Elastic Weight Consolidation (EWC) constrains parameters important to past tasks using a Fisher-based penalty \cite{kirkpatrick2017ewc}. iCaRL combines exemplar replay with representation learning for class-incremental settings \cite{rebuffi2017icarl}. Replay-based approaches are especially relevant in IoT, where limited buffers may still be feasible but must account for memory and energy budgets during their design \ cite {lourenco2025ondevice}.

\subsection{Contrastive Learning}
Contrastive learning optimizes embeddings so that positive pairs are close and negative pairs are far in the latent space~\cite{khosla2020supcon}. There are several contrastive learning-based techniques. SimCLR popularized the InfoNCE objective with strong augmentations \cite{chen2020simclr}. MoCo introduced a momentum encoder and queue for scalable negatives \cite{he2020moco}. BYOL demonstrated strong performance without explicit negatives in the online/target networks \cite{grill2020byol}. Supervised Contrastive Learning (SupCon) uses labels to define positives \cite{khosla2020supcon}.

\subsection{Contrastive Continual Learning}
Co$^2$L adapts supervised contrastive learning to continual settings via replay and a self-distillation mechanism that preserves instance relations \cite{cha2021co2l}. Later work explores buffer weighting and hard-negative mining (e.g., importance sampling) \cite{li2024importance}. Recent theory provides performance guarantees and motivates adaptive distillation coefficients (CILA) \cite{wen2024provable}. While much of the literature focuses on vision benchmarks, IoT introduces different modalities (tabular/time-series), drift patterns, and deployment constraints \cite{lourenco2025ondevice}.

\subsection{IoT and Edge Constraints}
TinyML and edge AI emphasize deploying machine learning in resource-constrained environments, such as IoT systems \cite{warden2019tinyml}. On-device edge learning for IoT data streams faces additional challenges: continuous arrival, limited storage, and evaluation beyond static accuracy (e.g., energy, latency, stability-plasticity) \cite{lourenco2025ondevice}. Additionally, IoT distributed learning paradigms such as Federated learning (FL), where data is processed locally for privacy, while coordinating model updates with a global curator \cite{mcmahan2017fedavg}, but concept drift across clients complicates convergence and stability \cite{mahdi2025fldrift}.

\section{Problem Formulation: CCL for IoT}
Consider an IoT stream segmented into increments $t=1,\dots, T$ (tasks, time windows, or drift regimes). At increment $t$, a learner receives data $\mathcal{D}_t = \{(x_i,y_i)\}$ (labels may be partial/absent) and updates parameters $\theta_t$ from $\theta_{t-1}$.

\subsection{Representation and Contrastive Views}
Let $f_\theta(\cdot)$ be an encoder producing embedding $z \in \mathbb{R}^d$. For an input $x$, define two stochastic transformations (augmentations) $a(\cdot)$ and $a'(\cdot)$ appropriate for IoT modality:
\[
z = f_\theta(a(x)), \quad z^+ = f_\theta(a'(x)).
\]
For time-series or tabular IoT data, augmentation methods may include jittering, scaling, masking, cropping, permutation, or frequency-domain perturbations (domain-specific).
\subsection{Contrastive Loss functions}
InfoNCE and Supervised Contrastive Learning (SupCon) are loss functions used in contrastive learning to develop representations by bringing similar samples closer together and distancing dissimilar ones. The key difference between them is in how they define "similar" samples (positives) and "dissimilar" samples (negatives). InfoNCE usually operates in a self-supervised manner, while SupCon relies on labeled data.

\subsubsection{InfoNCE Objective}
A standard contrastive loss for a minibatch is:
\begin{equation}
\mathcal{L}_{\text{NCE}} = - \sum_i \log
\frac{\exp(\mathrm{sim}(z_i,z_i^+)/\tau)}
{\sum_{j} \exp(\mathrm{sim}(z_i,z_j)/\tau)},
\end{equation}
where $\mathrm{sim}(\cdot,\cdot)$ is cosine similarity and $\tau$ is a temperature \cite{chen2020simclr,he2020moco}.

\subsubsection{Supervised Contrastive Loss}
With labels, SupCon defines positives as same-class samples:
\begin{equation}
\mathcal{L}_{\text{sup}} = \sum_i \frac{-1}{|P(i)|}
\sum_{p\in P(i)} \log \frac{\exp(\mathrm{sim}(z_i,z_p)/\tau)}
{\sum_{a\in A(i)} \exp(\mathrm{sim}(z_i,z_a)/\tau)},
\end{equation}
where $P(i)$ indexes positives and $A(i)$ all anchors in the batch \cite{khosla2020supcon}.

\subsection{Continual Objective with Replay and Distillation}
A common CCL objective combines: (i) contrastive loss on current data, (ii) contrastive/replay loss on buffered exemplars $\mathcal{M}$, and (iii) distillation preserving prior relations:
\begin{equation}
\min_\theta \ \mathcal{L}_{\text{CCL}} =
\mathcal{L}_{\text{ctr}}(\mathcal{D}_t \cup \mathcal{M})
+ \lambda \, \mathcal{L}_{\text{distill}}(\theta,\theta_{t-1}),
\end{equation}
as instantiated by Co$^2$L-style self-distillation \cite{cha2021co2l} and adaptive variants with theoretical grounding \cite{wen2024provable}.

\section{Method Taxonomy for IoT CCL}

In this section, we present a taxonomy of CCL techniques used in IoT, which is summarized in Table ~\ref{tab:taxonomy_ccl_iot}. This taxonomy is organized by CCL technique families, along with their concepts and properties. Additionally, Table ~\ref{tab:method_comparison} compares various continual learning approaches, both with and without contrastive learning, highlighting their characteristics in the IoT domain.

\begin{table*}[t]
\centering
\caption{Taxonomy of Contrastive Continual Learning (CCL) Methods for IoT Streams}
\label{tab:taxonomy_ccl_iot}
\begin{tabular}{@{}p{2.6cm} p{3.3cm} p{3.4cm} p{4.0cm} p{3.2cm}@{}}
\toprule
\textbf{Family} & \textbf{Core Idea} & \textbf{Typical Loss Components} & \textbf{IoT Strengths} & \textbf{Key Limitations} \\
\midrule
Replay-based CCL &
Store a small memory of past samples/embeddings &
$\mathcal{L}_{ctr}(\mathcal{D}_t \cup \mathcal{M})$; optional SupCon; buffer sampling strategies &
Strong forgetting mitigation; works with partial labels; simple to implement &
Memory/privacy constraints; buffer bias under non-IID streams \\
\addlinespace

Distillation-based CCL &
Match old vs. new representations or similarity structure &
$\mathcal{L}_{ctr}(\mathcal{D}_t) + \lambda \mathcal{L}_{distill}$ (logits/embeddings/relations) &
No raw data storage required; geometry preservation helps stability &
Requires teacher snapshot; sensitive to drift magnitude and $\lambda$ \\
\addlinespace

Regularization-based CCL &
Constrain important parameters while learning new data &
$\mathcal{L}_{ctr}(\mathcal{D}_t) + \beta \Omega(\theta,\theta_{t-1})$ &
Very low memory; good for small shifts and TinyML settings &
May under-adapt on strong drift; limited representation refresh \\
\addlinespace

Prototype/Exemplar CCL &
Keep class prototypes/cluster centroids instead of raw samples &
$\mathcal{L}_{ctr}$ with prototype positives/negatives; metric learning &
Compact memory; privacy-friendly; efficient on edge &
Prototype staleness under drift; hard for open-world classes \\
\addlinespace

Federated CCL &
Local CCL updates + global aggregation across clients &
Local: $\mathcal{L}_{ctr} + \mathcal{L}_{CL}$; Global: aggregation/align &
Privacy; multi-site learning; handles heterogeneity via representation alignment &
Client drift; communication cost; asynchronism and non-IID issues \\
\bottomrule
\end{tabular}
\end{table*}

\begin{table*}[t]
\centering
\caption{Comparison of Continual Learning Strategies for IoT (Practical View)}
\label{tab:method_comparison}
\begin{tabular}{@{}p{3.2cm} c c c c p{5.5cm}@{}}
\toprule
\textbf{Approach} & \textbf{Forgetting} & \textbf{Compute} & \textbf{Memory} & \textbf{Privacy Risk} & \textbf{Best Fit in IoT} \\
\midrule
Fine-tune (no CL) & High & Low & Low & Low & Only if drift is negligible or models are frequently retrained offline \\
\addlinespace
Regularization (e.g., EWC-like) & Med & Low & Low & Low & TinyML devices with limited RAM/flash; mild gradual drift \\
\addlinespace
Replay buffer (raw samples) & Low & Med & Med--High & Med--High & Edge gateways with storage; when labels exist and strong forgetting occurs \\
\addlinespace
Replay (embeddings/prototypes) & Low--Med & Low--Med & Low--Med & Low & Privacy-sensitive deployments; devices can store compact summaries \\
\addlinespace
CCL (Contrastive + Replay) & Very Low & Med--High & Med & Med & When representation quality matters (transfer/robustness), partial labels \\
\addlinespace
CCL (Contrastive + Distillation) & Low & Med & Low & Low & When storing data is not allowed; keep teacher snapshots instead \\
\addlinespace
Federated CCL & Low--Med & Med & Low & Low & Multi-site IoT; local data cannot leave device; heterogeneous clients \\
\bottomrule
\end{tabular}
\end{table*}

\subsection{Replay-Based Contrastive Continual Learning}
Replay-based contrastive continual learning maintains a limited memory buffer containing samples or representations from previous increments~\cite{lin2023pcr}. During training on new IoT data streams, the contrastive loss is computed jointly over current data and replayed samples. This mechanism explicitly reinforces previously learned invariances and class structure, effectively mitigating catastrophic forgetting. In IoT scenarios, replay is particularly valuable under sparse-labeling conditions, as contrastive objectives can leverage unlabeled historical data. However, replay buffers are constrained by device memory, energy, and privacy requirements, and naïve uniform sampling may introduce bias under non-IID or drifting streams.
Replay buffers store a small subset of prior samples $\mathcal{M}$ (or compressed prototypes). In IoT, storing raw data may be restricted (privacy) or not feasible (memory and processing constraints). CCL variants:
\begin{itemize}
  \item \textbf{Uniform replay:} sample past points uniformly.
  \item \textbf{Class-/client-balanced replay:} mitigate bias under non-IID streams.
  \item \textbf{Hard-negative aware replay:} prioritize informative negatives; importance sampling can improve representation separation \cite{li2024importance}.
\end{itemize}
Co$^2$L uses replay in combination with a supervised contrastive loss to maintain class structure \cite{cha2021co2l}.

\subsection{Distillation and Consistency Preservation}
Distillation-based CCL transfers knowledge from a previous model snapshot to an updated model by enforcing consistency across representations, logits, or similarity relations~\cite{zhang2023target, li2024continual}. Rather than storing raw IoT data, the new model learns to preserve embedding geometry or relational structure induced by the old model while optimizing a contrastive loss on incoming data. This approach is particularly attractive for privacy-sensitive IoT deployments where data retention is restricted. Nevertheless, distillation relies on the availability of a reliable teacher model and can struggle when concept drift is abrupt, as overly strong distillation may hinder necessary adaptation.
Distillation transfers knowledge from the previous model $\theta_{t-1}$ to the new model $\theta_t$:
\begin{equation}
\mathcal{L}_{\text{distill}} =
\sum_{x\in \mathcal{B}} \left\| g_{\theta}(x) - g_{\theta_{t-1}}(x) \right\|_2^2,
\end{equation}
where $g$ could be logits, embeddings, or relational similarities. In CCL, relation distillation (matching similarity matrices) helps preserve geometry across increments \cite{cha2021co2l}. Recent theory links final performance bounds to contrastive and distillation losses and suggests adaptive weighting (CILA) \cite{wen2024provable}.

\subsection{Regularization-Based CL as a Complement}
Regularization-based approaches constrain parameter updates that are deemed important in previous increments while allowing learning from new data~\cite{maschler2021regularization, ahn2019uncertainty, parisi2019continual, jung2020continual}. When combined with contrastive objectives, these methods encourage representation reuse while preventing destructive updates. In IoT contexts, regularization is computationally lightweight and well-suited to TinyML devices with severe memory constraints. However, such methods typically assume moderate distributional shifts and may underperform when sensor behavior or operating conditions change significantly, as the model’s capacity to reconfigure representations is limited.
Regularizers like EWC penalize parameter drift deemed important for older tasks \cite{kirkpatrick2017ewc}. In IoT, lightweight regularization can complement small replay buffers when memory is constrained, but may struggle under large shifts or changing label spaces.

\subsection{Prototype or Exemplar-Based Contrastive Continual Learning}

Prototype-based CCL replaces raw replay samples with compact class or cluster prototypes, such as centroids in embedding space~\cite{chen2024exemplar, zhu2021prototype}. Contrastive losses are then computed using these prototypes as positives or anchor samples, thereby significantly reducing memory usage. This design aligns well with IoT privacy and storage limitations and supports efficient edge-level learning. The main challenge arises under strong or recurring concept drift, where prototypes may become stale and fail to reflect current data distributions, particularly in open-world IoT settings with evolving classes.

\subsection{Federated Contrastive Continual Learning}

Federated CCL extends contrastive continual learning to distributed IoT environments in which data remain local to devices or sites~\cite{li2024continual}. Each client performs local contrastive continual updates, while a central coordinator aggregates model parameters or representation statistics. Contrastive objectives help align representations across heterogeneous clients, improving robustness to non-IID data. Despite its privacy advantages, federated CCL introduces new challenges, including asynchronous updates, client-specific drift, and increased communication overhead, which are particularly pronounced in large-scale IoT deployments.

\subsection{IoT-Specific Considerations: Tabular and Time-Series}

\begin{table*}[t]
\centering
\caption{Example Augmentations for Contrastive Learning in IoT Modalities}
\label{tab:iot_augmentations}
\begin{tabular}{@{}p{2.6cm} p{4.6cm} p{4.6cm} p{4.8cm}@{}}
\toprule
\textbf{Modality} & \textbf{Augmentations (Views)} & \textbf{Pros} & \textbf{Cons} \\
\midrule
Time-series sensors &
Jitter, scaling, time-warp, cropping, masking, permutation &
Preserve causal/local patterns; use light noise for robustness &
Over-warping distorts semantics; permutation may break temporal meaning \\
\addlinespace
Tabular telemetry &
Feature dropout/masking, mixup within class, noise injection, binning &
Use feature-aware perturbations; preserve constraints (ranges/units) &
Random shuffles violate feature semantics; label leakage via augmentation \\
\addlinespace
Audio (acoustic IoT) &
Time mask, freq mask, gain, background mixing &
Align with domain invariances; use spectrogram masking &
Aggressive mixing harms rare-event detection \\
\addlinespace
RF / CSI / WiFi sensing &
Phase jitter, subcarrier masking, temporal cropping &
Respect physical constraints; augment in the frequency domain carefully &
Augmentations that break channel structure can degrade downstream \\
\addlinespace
Vision (IoT cameras) &
Crop, color jitter, blur, flip (task-dependent) &
Follow established SSL practices & 
Augmentations that remove safety-critical cues (e.g., flip text/signals) \\
\bottomrule
\end{tabular}
\end{table*}
Various IoT data modalities are discussed in Table~\ref{tab:iot_augmentations}.

Most CCL benchmarks are vision-centric, whereas IoT often relies on tabular telemetry and time-series data. Open-world and out-of-distribution behavior is common. Recent work highlights that on-device IoT streams require careful treatment of streaming protocols and resource-aware evaluation \cite{lourenco2025ondevice}. Emerging tabular continual contrastive learning directions also indicate growing attention to non-vision modalities \cite{ginanjar2025tccl}.

\section{IoT CCL Architecture}

\begin{table*}[t]
\centering
\caption{Deployment Trade-offs for Contrastive Continual Learning in IoT}
\label{tab:deployment_tradeoffs}
\begin{tabular}{@{}p{2.2cm} p{3.0cm} p{3.1cm} p{3.2cm} p{4.9cm}@{}}
\toprule
\textbf{Tier} & \textbf{Typical Resources} & \textbf{What to Train} & \textbf{Memory Strategy} & \textbf{Recommended CCL Choices} \\
\midrule
Device (TinyML) &
Very low RAM/flash; strict energy &
Light adapter/prompt, last-layer, or periodic encoder updates &
Prototypes/quantized embeddings; tiny buffer &
Regularization + light contrastive; prototype replay; infrequent updates \\
\addlinespace
Edge/Gateway &
Moderate compute/storage; local aggregation &
Encoder + head; SSL pretraining + incremental fine-tune &
Replay buffer (raw or compressed); drift-aware sampling &
Contrastive + replay; relation distillation; drift-triggered updates \\
\addlinespace
Cloud/Coordinator &
High compute; cross-site visibility &
Global consolidation, validation, distillation, federation coordinator &
Central model snapshots; optional synthetic replay &
Federated CCL; global distillation; auditing/rollback \\
\bottomrule
\end{tabular}
\end{table*}

\begin{figure}[t]
\centering
\includegraphics[width=\linewidth]{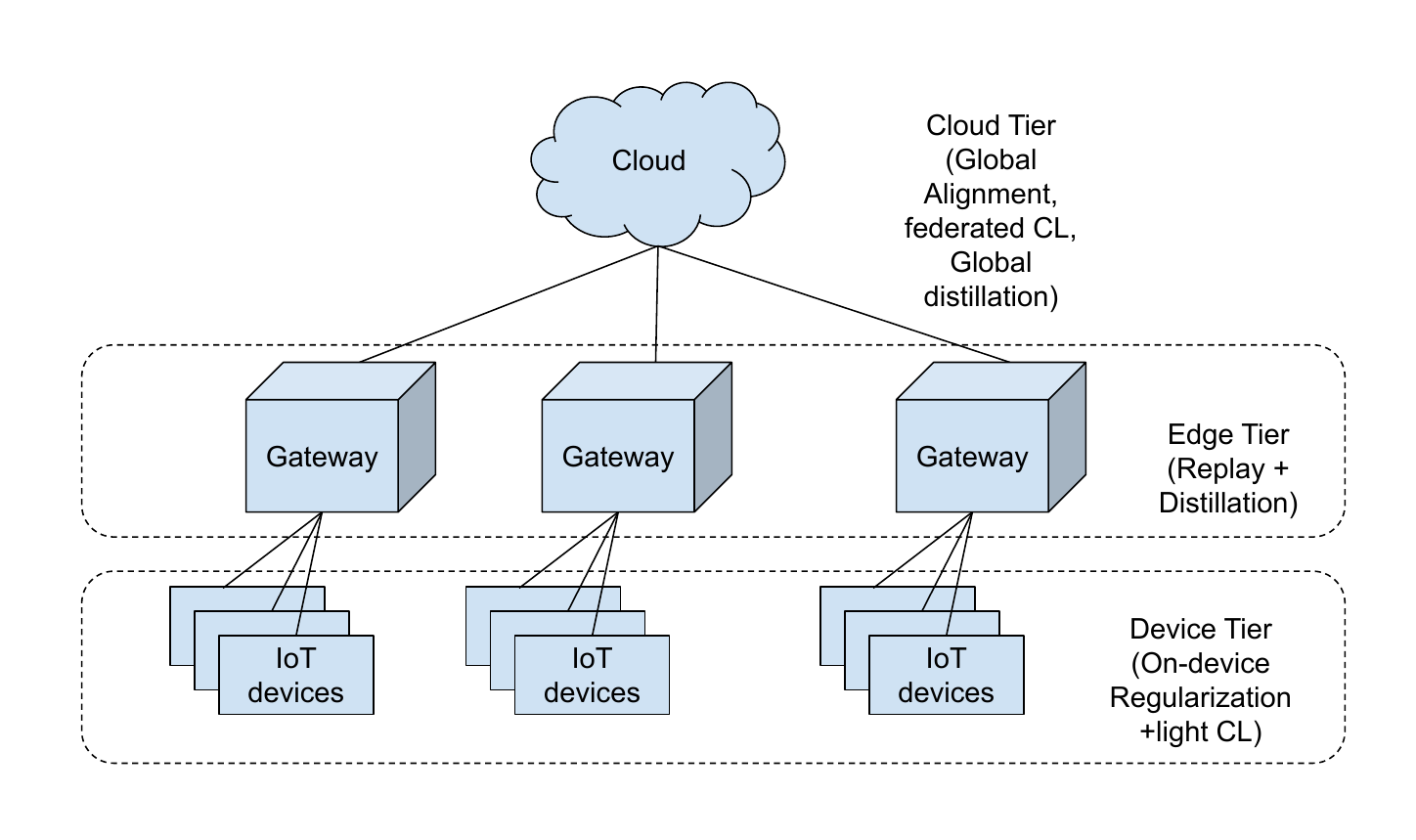}
\caption{Three-tier architecture for IoT CCL across device, edge, and cloud.}
\label{fig:arch}
\end{figure}

Figure~\ref{fig:arch} depicts a practical three-tier IoT CCL deployment architecture.

Additionally, we provide a summary of the three-tier deployment architecture of IoT CCL and its trade-offs with respect to resource requirements, and discuss various recommendations for each CCL choice in Table ~\ref{tab:deployment_tradeoffs}. Below, we examine each tier in the architecture.

\subsection{Tier 1: Device (TinyML) Layer}
IoT devices, such as those used in human activity recognition or security monitoring systems, collect streams of data from their sensors and may run inference continuously~\cite{chathoth2024dynamic, xu2024drift_iot}. Training is optional and typically constrained:
\begin{itemize}
  \item Maintain a tiny buffer of embeddings/prototypes (rather than raw samples) when possible.
  \item Use cheap augmentations and small encoders.
  \item Schedule updates opportunistically (charging, low-duty cycles).
\end{itemize}
TinyML practices emphasize extreme memory and power budgets \cite{warden2019tinyml}.

\subsection{Tier 2: Edge/Gateway Layer}
Gateways play a major role in IoT systems by offloading data processing and protocol translation and enabling external connectivity. Gateways can also perform heavier on-site training and aggregation:
\begin{itemize}
  \item Store a larger replay buffer with privacy controls.
  \item Run contrastive pretraining on unlabeled local data; fine-tune with sparse labels.
  \item Detect drift and trigger incremental updates.
\end{itemize}
On-device/edge learning surveys emphasize multi-criteria evaluation and stream-aware design \cite{lourenco2025ondevice}.

\subsection{Tier 3: Cloud/Coordinator Layer}
The cloud is integral to modern IoT devices, providing intelligent, data-driven services and insights for users and devices. The cloud-based service provider can coordinate FL or periodic consolidation from each client to facilitate training in a privacy-preserving manner:
\begin{itemize}
  \item Aggregate updates across sites/clients via FedAvg-style methods \cite{mcmahan2017fedavg}.
  \item Handle heterogeneous drift and asynchronous updates, which are central challenges in FL under drift \cite{mahdi2025fldrift}.
  \item Provide model versioning, validation, and rollback safeguards.
\end{itemize}
\section{Evaluation Protocols and Metrics for IoT CCL}
This section discusses evaluation protocols and metrics to consider for CCl in an IoT setting.
\subsection{Protocols}
\textbf{Stream-incremental:} update at fixed time windows (e.g., hourly/daily) or drift-triggered intervals. This is based on the observation that IoT device data often consists of time-series data streams configured by the IoT device administrator.

\textbf{Task-incremental:} explicit regime segmentation (site changes, firmware versions). This occurs when an IoT device is decommissioned and then recommissioned at a new site, or when its firmware is upgraded to include additional features.

\textbf{Label regimes:} fully labeled, partially labeled, delayed labels, or self-supervised only. This is based on the observation that IoT devices generate large volumes of data, and the rate at which they are generated is difficult to characterize due to their limited processing power and resource constraints.

\subsection{Accuracy and Forgetting Metrics}

\begin{table*}[t]
\centering
\caption{Recommended Metrics for IoT Contrastive Continual Learning}
\label{tab:metrics}
\begin{tabular}{@{}p{3.4cm} p{12.2cm}@{}}
\toprule
\textbf{Metric} & \textbf{Definition} \\
\midrule
Avg. Accuracy &
$\frac{1}{T}\sum_{k=1}^{T} A_{T,k}$; final performance across all increments \\
\addlinespace
Forgetting &
$\frac{1}{T-1}\sum_{k=1}^{T-1}\left(\max_{t \in [k,T]} A_{t,k} - A_{T,k}\right)$ \\
\addlinespace
Forward Transfer &
Performance on increment $t$ before training on $t$ (representation reuse) \\
\addlinespace
Stability--Plasticity &
Report both retention (stability) and new-task gain (plasticity), e.g., $\Delta A_{t,t}$ vs. forgetting \\
\addlinespace
Buffer Cost &
$\lvert \mathcal{M} \rvert$, bytes stored, and replay sampling rate \\
\addlinespace
Compute \& Energy &
Update time (ms), MACs, and energy per update (mJ) for device/edge tiers \\
\addlinespace
Communication (FL) &
Bytes/round, rounds to reach target accuracy, dropout robustness \\
\bottomrule
\end{tabular}
\end{table*}
Table ~\ref{tab:metrics} summarizes the metrics recommended for measuring the performance of CCL in IIoT.
Let $A_{t,k}$ be the accuracy on the test set of increment $k$ after training up to $t$.
\begin{itemize}
  \item \textbf{Average accuracy:} $\frac{1}{T}\sum_{k=1}^T A_{T,k}$\\
  \item \textbf{Forgetting:} $\frac{1}{T-1}\sum_{k=1}^{T-1}\left(\max_{t\in\{k,\dots,T\}} A_{t,k} - A_{T,k}\right)$\\
  \item \textbf{Forward transfer:} performance on unseen increments before training them.
\end{itemize}

\subsection{Resource Metrics (Crucial for IoT)}
Report:
\begin{itemize}
  \item \textbf{Peak RAM} and \textbf{flash/storage} footprint (buffer + optimizer state).\\
  \item \textbf{Energy per update} and \textbf{time-to-update}.\\
  \item \textbf{Bandwidth} in federated settings (bytes/round, rounds to target).
\end{itemize}
IoT edge-learning surveys recommend multi-criteria metrics beyond accuracy \cite{lourenco2025ondevice}.

\section{Open Challenges and Research Directions}
Despite the advantages of CCL in improving a model's utility in IoT, several challenges remain.
\subsection{Concept Drift and Open-World IoT}
IoT deployments face recurring and abrupt drift. Robust drift detection and adaptation loops are needed, especially for anomaly detection and security monitoring under evolving patterns \cite{xu2024drift_iot}. A key gap is the lack of standardized CCL benchmarks for realistic IoT drift.

\subsection{Tabular/Time-Series Contrastive Objectives}
Defining augmentations and positives for tabular telemetry is nontrivial; naive corruptions can destroy semantics. Tabular CCL frameworks are emerging but require further study to address IoT-specific constraints and out-of-distribution behavior \cite{ginanjar2025tccl}.

\subsection{Federated Continual Contrastive Learning Under Drift}
FL under drift creates client-dependent, asynchronous, and non-stationary. Systematic reviews emphasize open issues in drift detection timing, aggregation under heterogeneous drift, and stability \cite{mahdi2025fldrift}. Contrastive objectives can help align representations across clients but require careful design of privacy and communication.

\subsection{Theory-Guided and Adaptive Weighting}
Theory-backed methods (e.g., adaptive distillation coefficients) are promising for principled hyperparameter selection and stability guarantees \cite{wen2024provable}. Translating these ideas to IoT modalities and resource limits remains open.

\subsection{Energy-Aware and Safety-Critical Updating}
Edge updates can be risky (model regressions) and costly (battery consumption). Research is needed on safe deployment mechanisms, including shadow evaluation, rollback policies, and uncertainty-aware update triggers.

\section{Conclusion}
Contrastive continual learning offers a compelling path for IoT systems that must adapt over time while remaining robust, secure, privacy-preserving, and resource-efficient. By unifying objectives such as contrastive learning, replay, and distillation, and aligning them with IoT device architectures, edge and cloud architectures, and evaluation practices, we can move toward a practical, private, and secure, continually learning IoT. Future work in this direction could focus on prioritizing realistic drift benchmarks for IoT modalities, federated drift robustness, and energy-aware, safe updating.

\bibliographystyle{IEEEtran}
\bibliography{bib}

\end{document}